# From Process Models to Composable Agentic Workflows: Combining BPMN, Agents and Enterprise Knowledge

Christian Fillies[1], Jan Fillies[2], Frauke Weichhardt[1], Gay Anne Wood-Albrecht[1]
Semtation GmbH[1], FU Berlin[2]
Correspondence to: Christian Fillies, `cfillies@semtalk.com`

**Abstract**

Large Language Model (LLM)-based agents can reason about tasks, use tools and collaborate with other agents. However, as agentic applications grow in complexity, their coordination can become difficult to understand, monitor and improve. Workflows become implicit because decisions are distributed across prompts, tool calls and application logic. Recursive Capability Composition is introduced in which agents can complete business processes and then become independent Agentic Tools with external capabilities.

This paper presents a BPMN-based approach to Agentic Workflows that combines explicit process orchestration with autonomous reasoning. SemTalk BPMN models are transformed into executable workflows using a LangGraph / Pregel runtime. Agentic Tasks introduce LLM-based reasoning at selected activities, while conventional BPMN retains responsibility for process flows, deterministic decisions, system interactions, and human approval processes. Agentic Participants provide reusable capability interfaces that can encapsulate agent behavior or invoke other executable processes.

We argue that BPMN-based orchestration offers a practical foundation for making multi-agent systems more understandable and operationally manageable. Integration of agent and tool execution, as they apply to explicit process models, allows developers and operators to have a shared view of workflow structures and execution progress. This approach preserves the flexibility of LLM-based reasoning while creating a more efficient platform for debugging, monitoring, and systematic optimization.

Optimal integration of Enterprise AI agents requires an organizationally specific context. Retrieval-Augmented Generation provides access to enterprise information, but this knowledge is not inherently integrated into general-purpose LLMs. Process Models and Knowledge Graphs provide a structured source of context knowledge for agents. They describe activities and their ordering, as well as participants, information objects, decision branches, and relationships between processes. BPMN integration contains the information-dependent progression of activities in a process orchestration context, and it describes how an organization performs its business functions. This provides agents with the organization-specific contextual knowledge needed to support agent decisions and actions.

The contribution of our work is threefold:

- A Model-Driven Approach is presented in which BPMN Process Models are transformed into Executable Agent Workflows retain conventional BPMN to provide explicit process orchestration.
- A Recursive Capability Composition is introduced in which agents can be tools that complete Business Processes themselves that can become tools with external capabilities.
- Process Models and Ontologies are used to provide organization-specific context.

# 1. Introduction

Large Language Models (LLMs) have changed how software systems are built and used. Users can formulate requests in natural language, while developers can use prompting and AI-based tools to create increasingly complex applications. The emerging practice of “vibe coding” illustrates this shift [8]. LLM-based agents extend it further by selecting tools, interacting with other agents, and carrying out tasks with a degree of autonomy.

However, as Agentic Systems grow in complexity, they become harder to understand, maintain, monitor and improve. Decisions may be distributed across prompts, tool calls and application logic, making it difficult to reconstruct why a system took a particular path or to determine where an error occurred. An important consideration in Agent Orchestration is to make workflow structures, decisions and states more explicit [8] [10][11]. Workflow automation platforms such as n8n and Make, as well as agent frameworks such as LangChain [3], provide different mechanisms for defining and coordinating tasks. Business Process Model and Notation (BPMN) [2], traditionally used by workflow engines such as Camunda, offers an established notation for representing activities, decisions and the flow of work.

SemTalk [12] provides a Model-Driven Runtime in which BPMN Process Models are transformed into Executable Agentic Workflows. The BPMN Model defines orchestration, while individual activities can be performed by agents where flexible reasoning is required. This separation allows existing process models to be extended with agentic capabilities without replacing their explicit control flow. It also provides a common representation that can be used to understand the intended workflow and relate it to its execution.

Organizations are increasingly exploring how agentic technologies can improve existing business processes. BPMN Process Models can provide an important foundation for this transition because they make the relationship between agentic and conventional activities explicit. A process can combine LLM-based reasoning with deterministic decisions, system tasks and human tasks while retaining a comprehensible representation of how work progresses. An agent can therefore perform a particular activity without requiring the entire process to be delegated to an autonomous agent.

Enterprise agents need to understand not only general concepts, but also how a particular organization operates. General-purpose LLMs do not inherently contain an organization's specific processes, responsibilities, rules, and operational knowledge. Retrieval-Augmented Generation (RAG), including approaches used in systems such as Microsoft Copilot, can make enterprise information available at runtime. However, the quality and relevance of the resulting context depend on the information that can be retrieved and how it is represented.

Process Models and Knowledge Graphs can provide this enterprise-specific context in a structured form. Process Models play two complementary roles. First, they describe how work is organized by describing which activities are performed, in what order, and along which possible paths a process may proceed. This makes them a basis for explicit process orchestration. Second, they capture how an organization performs a particular business function, providing agents with knowledge about activities, responsibilities, and relationships within that function. A Process Model can therefore serve both as an executable specification of work and as a source of organizational context for agent reasoning.

Composability is a further aspect of this approach. Modern agent architectures commonly allow agents to be exposed as tools. SemTalk extends this idea through Agentic Participants, which provide reusable capability interfaces. An agent can invoke such a capability without needing to know its internal implementation. The capability may involve direct agent reasoning or be backed by a BPMN Process that contains agents, human tasks, system tasks and subprocesses.

The key architectural insight is not just that a BPMN Process can be wrapped as a tool, it is that BPMN Participants provide a uniform capability boundary between Process Orchestration and Agentic Reasoning. An Agentic Participant is exposed as a callable tool and may encapsulate LLM, tool, retrieval, and contextual knowledge. If the participant is associated with a BPMN Process, the same tool interface can invoke the executable representation of that process.

By combining explicit Process Orchestration with composable agent capabilities, BPMN-based Agentic Workflows make multi-agent systems easier to understand, monitor, debug and optimize. Process Models and Knowledge Graphs can additionally provide organizational context for agents, but the primary focus of this paper is to describe the use of BPMN as an executable structure for coordinating agentic and conventional work.

# 2. Business Process Modeling and Agentic Systems

Business Process Model and Notation (BPMN) is a standardized notation for describing Business Processes. Its constructs include Activities, Events, Gateways, Participants, Data Objects and Message Flows. BPMN supports process documentation, communication and analysis, and can also serve as the basis for process execution.

BPMN provides an explicit representation of organizational work. It describes which activities are performed, how they are connected, which paths a process may follow, and which participants are involved. This structure distinguishes process orchestration from the implementation of individual activities. BPMN defines how work progresses, while the execution of an activity may be provided by different types of software components.

An LLM-based agent can be understood as a software component that combines language-based understanding and reasoning with access to tools or other resources [4][5]. Agentic Workflows connect such components into larger structures. Frameworks such as LangGraph provide executable graph abstractions in which computational steps, agents, tools and transitions form a Runtime Workflow. The relationship to BPMN is therefore twofold: BPMN can provide an explicit model of the work to be performed, while Agent Runtime provides mechanisms for executing agentic behavior within that structure.

Recent research has investigated how BPMN can be extended to represent Human-Agentic Collaborative Workflows. Ait et al. [1] propose extensions that are motivated by limitations of standard BPMN in representing concepts such as agent roles, trust, uncertainty, reflection and collaboration strategies. Their approach introduces constructs such as “AgenticLanes” and “AgenticTasks” and extends collaboration and merges semantics. This work addresses how agent-specific behavior and collaboration concepts can be represented explicitly in BPMN.

# 3. From BPMN to Executable Agentic Workflows

SemTalk transforms BPMN Process Models into executable Agentic Workflows by mapping BPMN elements to runtime operations in a LangGraph/Pregel execution environment. The process model is not merely displayed as a graph: its activities, sequence flows, gateways and conditions determine which operations are executed and how control proceeds through the workflow. The runtime therefore preserves the explicit structure of the BPMN process while allowing individual activities to be implemented by different types of components, including LLM-based agents.

## 3.1 Combining Conventional and Agentic Activities

SemTalk allows an agent to be associated with a process activity. This association does not change the overall structure of the process. Instead, it determines how this activity is executed.

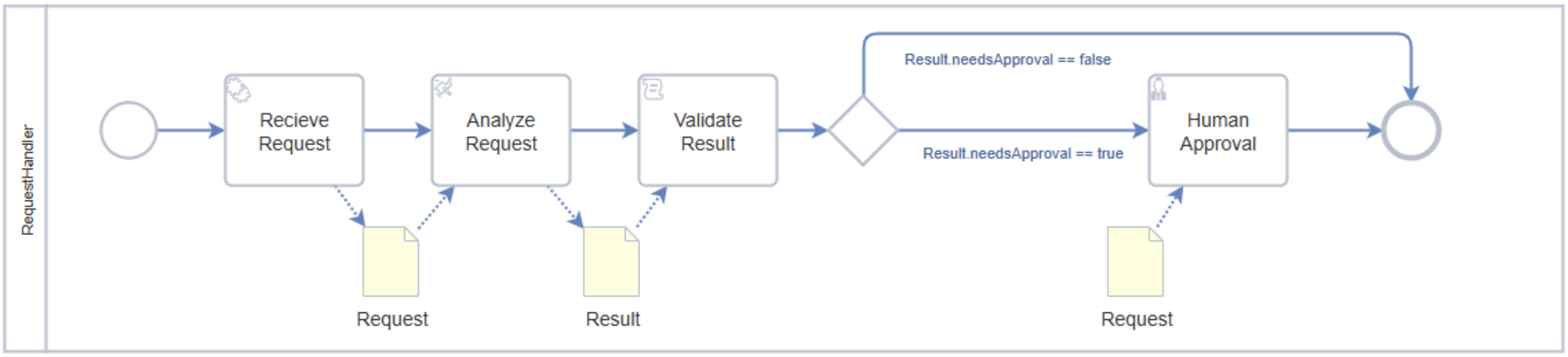

*Figure 1: BPMN Process with sequential Activities and a Gateway*

Figure 1 illustrates a BPMN Process with sequential activities, a gateway that determines the subsequent path, and data objects associated with individual activities. When transformed into an executable workflow, the BPMN structure determines the order of execution and the conditions for taking alternative paths. Activities operate on process data, while a gateway controls how execution proceeds. In the runtime, this structure is combined with a shared workflow state, allowing the results of one activity to influence subsequent execution. Individual activities can be implemented as conventional tasks or delegated to agents without changing the overall process structure.

In the example, the *Analyze Request* Activity is performed by an LLM-based agent. This is one illustration of a general principle: any activity that benefits from autonomous reasoning can be assigned to an agent, while activities that do not require such reasoning can retain their conventional implementations. A process may therefore combine Agentic Activities with Script Tasks, Service Tasks and User Tasks without replacing its existing BPMN structure.

The Process Model remains responsible for defining the sequence of activities, the conditions under which alternative paths are taken, and the points at which subsequent work can proceed. The agent, in contrast, determines how its assigned activity is performed. This separation allows explicit process orchestration and flexible agent reasoning to coexist within the same workflow.

## 3.2 From BPMN Elements to Runtime Execution

The SemTalk Workflow Runtime transforms the Process Model into an executable LangGraph Graph. BPMN elements are mapped to runtime nodes, and sequence flows determine the possible transitions between them. Where a gateway controls the flow, its conditions determine which transition is taken. The graph is thus executed according to the semantics of the Process Model rather than being an arbitrary network of computational steps.

| **BPMN 2.0 Element** | **LangGraph / Pregel Mapping** | **Runtime Execution Mechanics** |
|---|---|---|
| **Data Object / Property** | State Schema (Typed Dict) | Acts as the thread-safe, centralized global state containing process variables. |
| **Service / Script Task** | Deterministic Node | Instead of python/application code we modify the State directly without human or LLM intervention using assignments, avoiding risky code. |
| **Agentic Task** | LLM-backed Node (with Agent Loop) | Spatially isolates the LLM; hooks into an iterative loop of prompting, tool-calling, and reasoning until a __end__ token or objective condition is reached. |
| **Sequence Flow** | Edge / Conditional Edge | Establishes execution pathways. Evaluates the state to route the execution token from one node to the next. |
| **Exclusive Gateway (XOR)** | Routing Function (Conditional Edge) | Computes a deterministic boolean or string expression against current state variables to select a single outbound edge. |
| **Parallel Gateway (AND)** | Diverging / Converging Branching | **Fork:** Emits multiple state updates simultaneously. **Join:** Pregel superstep barrier that stalls execution until all incoming parallel tokens are received. |
| **Agentic Participant / Pool** | Encapsulated Subgraph + Tool Schema | Exposed to external agents via the **Model Context Protocol (MCP)** or standard JSON tool schemas. Intercepting a tool call spins up an isolated child Pregel graph. |

The execution of each activity depends on its BPMN type. Script Tasks execute deterministic logic and manipulate process variables; Service Tasks invoke services; and User Tasks support human interaction through dialogs, including Adaptive Cards. Agentic Tasks delegate the activity to an LLM-based agent. The runtime coordinates these different execution types within one workflow, so that the output of an activity becomes available to subsequent activities through the shared workflow state.

Process variables connect the individual steps. Activities can read inputs from the workflow state and write results back as variable values. Gateways can then evaluate conditions over this state to determine how the process continues. This combination of explicit control flow, shared state, typed activity execution and conditional routing makes the BPMN-derived graph executable.

The same structure also provides a basis for implementing guardrails and harnesses around agent behavior. The process can constrain which activities are reached, which conditions determine the next step, and which state is passed between activities. Deterministic logic can remain outside the agent, rather than being delegated to an LLM unnecessarily. LangGraph interrupts can support human interaction and execution debugging, allowing execution to pause at designated points instead of requiring the entire Workflow to run without intervention.

## 3.3 Agentic Tasks as an Execution Type

An Agentic Task is a BPMN Activity whose internal behavior is delegated to an LLM-based agent. Its configuration may specify a prompt, an LLM and its parameters, and the MCP Tools available to the agent. Unlike a Script Task, which executes specified logic, or a Service Task, which invokes a defined service, an Agentic Task leaves the internal reasoning strategy to the agent.

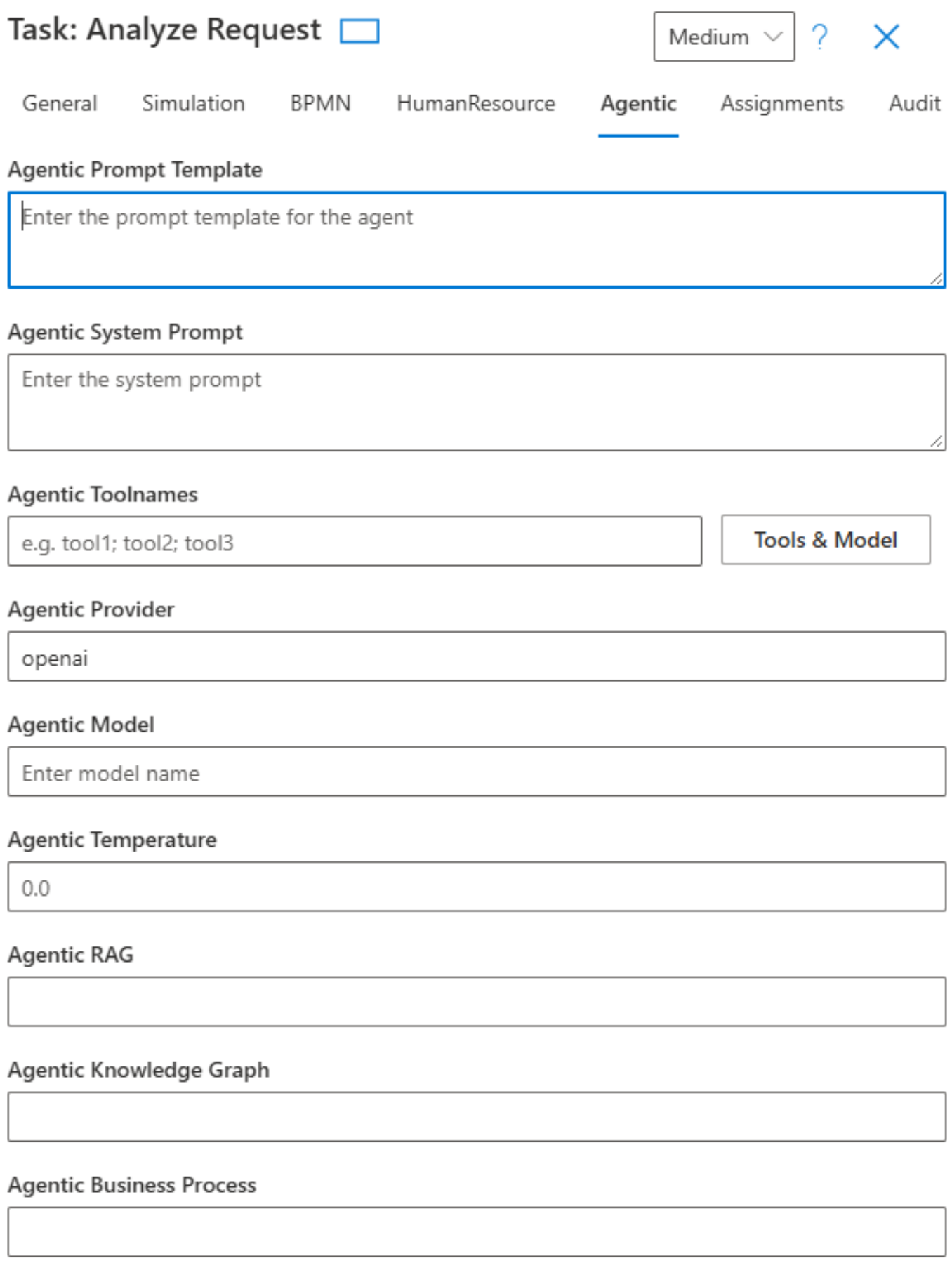

*Figure 2: Properties of an Agentic Task*

Agents receive inputs from the workflow state and return outputs that become part of the same state. During execution, agents may use MCP Tools, retrieve relevant information, access process knowledge, or they may interact with other agents, depending on their configuration and available capabilities (Figure 2). The surrounding BPMN Process continues to determine when the activity is invoked, and how it influences subsequent execution.

This distinction is central to the approach: BPMN determines when an activity is executed and how the workflow proceeds, while the agent determines how the assigned work is performed. Agentic Tasks are therefore not a separate workflow language or an alternative to BPMN orchestration. They provide an additional way to execute an activity within an existing BPMN Process.

Multiple Agentic Tasks can use different agents, prompts, tools and retrieval sources while remaining integrated with conventional activities and gateway logic. Agentic capabilities can consequently be introduced incrementally, at the points where they add value, without requiring the entire process to become autonomous.

## 3.3 Agentic Participants as Tools

Following LangGraph we call in SemTalk a participant marked as "Agentic" an "Agentic Participant". It is represented in the LangGraph runtime as an autonomous agent and, at the same time, as a callable tool. This tool provides an interface through which other agents or Agentic Tasks can request the capabilities associated with that participant.

An Agentic Participant can encapsulate several forms of capability such as an LLM and associated system prompt, tools available to the agent, retrieval or RAG capabilities, semantic or process context, other Agentic Participants exposed as tools and, importantly, an associated BPMN process.

This leads to an important distinction between an Agentic Task and an Agentic Participant:

- An Agentic Task is an activity in the current process that requires agentic reasoning.
- An Agentic Participant is a reusable agent capability that can be invoked by such a task or by another agent.

An Agentic Participant does not necessarily have an own BPMN process. If it does not, invocation of the participant tool results in the agent performing its capability using its configured LLM, tools, context and instructions. If it does have an associated BPMN process, however, the meaning of the tool invocation changes. The tool becomes an entry point into that process, and the runtime invokes the corresponding LangGraph representation of the business process. This creates a particularly useful form of composition. From the perspective of the calling agent, both cases have the same tool interface. Internally, however, the capability may range from a relatively simple agent invocation to an entire executable business process. The calling agent therefore does not need to know whether the participant is implemented by a single agentic operation or by a BPMN workflow containing multiple activities, other agents, human tasks and system services.

This is a stronger form of encapsulation than simply exposing a process as a tool. The process is not itself the primary abstraction exposed to the agent. The Agentic Participant is. The process becomes the implementation of that participant's capability when one is defined.

## 3.4 Recursive Composition

The Participant-as-Tool mechanism provides the basis for Recursive Composition.

An Agentic Task may invoke an Agentic Participant. That participant may use its own tools, including other Agentic Participants. If the participant has an associated BPMN Process, the invocation may instead enter that process, whose execution can contain further agentic tasks and participants.

The resulting structure is recursive: Processes can invoke agents, agents can invoke other agents, and agents can invoke business processes through Agentic Participants.

The important point is that Agentic Participants are not different mechanisms stitched together at the application level. They share the same execution environment. A tool call can therefore switch from agent reasoning into BPMN execution and back again.

It is not simply that processes can be tools, it means that BPMN Participants can represent reusable agent capabilities, and those capabilities may themselves be implemented by executable BPMN Processes.

This also explains why the approach naturally supports recursive agentic workflows. The boundary exposed to an agent is a capability - the implementation behind that capability can range from a single LLM-based action to a complete business process.

### 3.5. Externalizing Process Capabilities

Once an agent and its processes are represented as a capability, the same interface can be used for functions beyond local runtime.

The Model Context Protocol (MCP) [6], [7] provides a standardized mechanism for applications and agents to discover and invoke tools Model Contect Protocol. A Process Repository can be implemented as an MCP server which exposes tools which are implemented by any optionally agentic business process.

This allows an external agent to invoke enterprise capabilities that have been modeled without having to understand its internal process structure. The same architecture can also work in the other direction. External tools and agents can be made available to a process. In this approach MCP is not the primary contributor, it simply provides an interface through which modeled capabilities can participate in a larger agent ecosystem.

## 4. Process Models as Enterprise-Specific Context

Enterprise-specific context for agents can be roughly separated into 3 major layers as depicted in Figure 3.

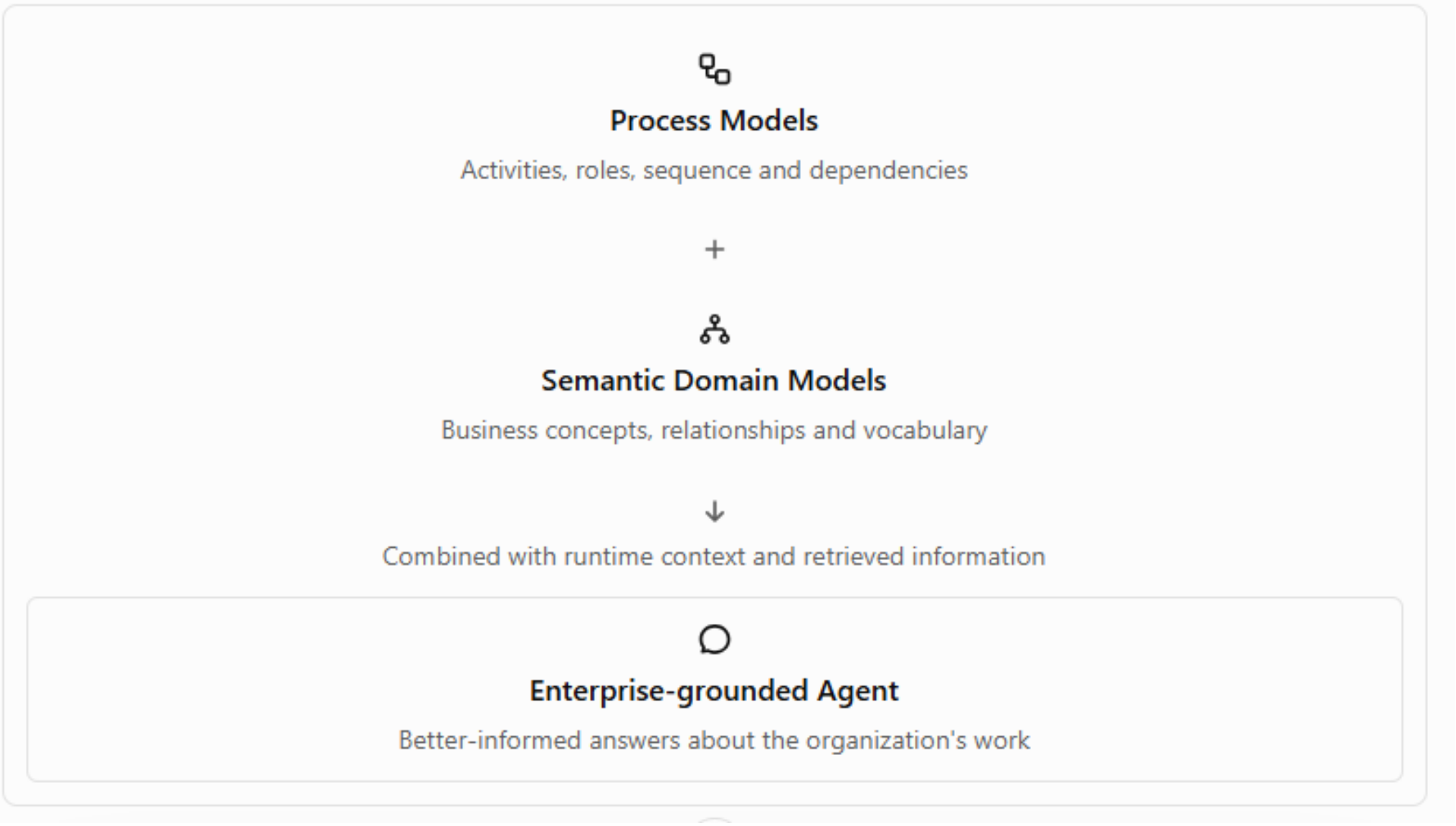

*Figure 3: Enterprise specific context for Agents*

## 4.1 From Process Description to Organizational Knowledge

Process Models contain more than descriptions of activity sequences. They also capture organizational knowledge about participants, responsibilities, business objects, decision points and relationships between processes. This information can help agents interpret business requests in the context of how a particular organization operates.

A process model can therefore serve two complementary purposes: as a source of knowledge to be retrieved and interpreted by an agent, and as an executable instruction that determines what should happen next. The same model can support both uses, connecting an agent's understanding of organizational work with the execution of that work.

## 4.2 Semantic Grounding

SemTalk connects Process Modeling with Semantic Domain Modeling. Business Objects and other process elements can be linked to organizational vocabularies and knowledge models, providing agents with context beyond the names and descriptions of individual activities.

Consider an activity called Handle Quality Return. Its meaning depends on the organization's business domain. It might involve a delivery, a return delivery, a customer, a material, a quality inspection, and subsequent logistical or financial activities. Semantic Models provide the vocabulary and relationships needed to interpret these concepts in their organizational context.

The resulting agent context can combine three complementary sources:

- Process Context, describing the current activity, its participants, inputs, outputs and related processes;
- Domain Context, describing Business Objects, relationships and organizational terminology; and
- Runtime Context, containing the current state, conversation, process variables and Tool results.

Unlike a conventional text-based retrieval approach that treats process documentation primarily as text, this approach can preserve and expose the structured relationships represented in Process Models and Knowledge Models. These relationships provide agents with enterprise-specific context for interpreting requests and performing their assigned tasks.

### 4.3 Process Memory and Organizational Context

Enterprise Process Repositories preserve organizational knowledge accumulated over time. Alongside current processes, they may contain Process Variants, Reference Processes, legacy descriptions and knowledge about business practices that remain relevant even after the underlying IT systems have changed. We use the term Process Memory for this accumulated knowledge. It can provide agents with enterprise-specific information that is unlikely to be present in the general-purpose LLM's pretrained knowledge.

The context available to an agent can range from the current activity with its immediate participants, inputs and outputs to surrounding activities, subprocesses and alternative paths, and to other related processes, departments, stakeholders and business capabilities across the enterprise. This broader context helps to connect an individual activity to the organizational setting in which it takes place.

The executable process tells the agent what it is doing. The broader process landscape helps it to understand where that activity fits in the organization. By grounding responses in relevant process memory and relationships, agents can produce better-informed answers to questions that cannot be answered from a single activity alone.

## 5. Related Work

Research into Agentic Business Process Management is developing rapidly, particularly along the boundaries of explicit process control and autonomous reasoning [4]. Existing literature can be structured into three distinct branches.

The first branch of literature focuses on extending traditional modeling languages to capture agent-specific behaviors. Ait et al. [1] propose a formal BPMN extension targeting Human-Agentic Collaborative Workflows, introducing primitives for agent profiles, trust metrics, uncertainty, and reflection mechanisms. While their approach successfully provides a dedicated modeling environment for agent interactions, it primarily serves as a design-time specification rather than an integrated model-driven runtime environment.

The second branch addresses execution through Agent Orchestration Frameworks and graph-based runtimes, such as LangGraph and Make. These systems construct highly dynamic workflows by composing agents and tool-calling interfaces. Ait et al. [1] identify these execution frameworks as essential orchestration technologies. However, as noted in our introduction, purely code or prompt-driven orchestration leaves the global workflow implicit, distributed, and difficult to manage at an enterprise scale.

The third branch combines process models with semantic web technologies, ontology frameworks, and Enterprise Knowledge Graphs. This approach recognizes that business process models contain deeply structured organizational information that can be linked to broader enterprise knowledge bases. While these systems excel at contextual tracking, they frequently isolate process data as a static knowledge source rather than an active execution driver.

The SemTalk approach presented in this paper uniquely synthesizes these three distinct perspectives by leveraging the dual role of process models. Unlike frameworks that treat process models strictly as deterministic workflow specifications, or as independent notation libraries, SemTalk treats BPMN models simultaneously as:

1. Rigid specifications for executable LangGraph/Pregel Workflows,

2. Dynamic sources of structured organizational knowledge for LLM-reasoning contexts, and as,

3. Clear boundaries for recursive, composable agentic capabilities.

By implementing this semantic layer, our approach explicitly differentiates between the top-level process orchestration and the encapsulated autonomous agent behavior by directly resolving the structural visibility issues inherent in modern multi-agent systems.

# 6. Limitations and Future Work

Our current implementation demonstrates the feasibility of model-driven Agentic Workflows, but systematic evaluation at larger scale remains future work. In particular, the preservation of BPMN execution semantics must be assessed across representative processes, including conditional routing, parallel paths, state propagation and human interaction. Formal validation of the BPMN-to-LangGraph transformation would provide a stronger basis for establishing execution correctness.

A second limitation concerns the boundary between deterministic process control and probabilistic agent behavior. Explicit BPMN structures constrain when an agent is invoked and how execution proceeds, but it cannot guarantee that the agent produces a correct result or selects an appropriate tool. Agent-specific properties such as uncertainty, trust and reflection are not yet first-class concepts in the SemTalk process model. Their explicit representation may become necessary for workflows that require stronger governance.

Composing Agentic Participants and exposing them as callable capabilities also introduces requirements for authorization, auditability and controlled execution, particularly when a capability can initiate a process involving business systems or human actions. In addition, the quality of enterprise-specific context depends on the completeness of Process Models, Semantic Domain Models and retrieval mechanisms.

Future work will therefore focus on validating BPMN-to-runtime transformations, evaluating the execution of increasingly complex multi-agent workflows, and improving execution traces so that Agent and Tool invocations can be inspected and analyzed. Further directions include generating capability interfaces from Process Models and investigating how execution traces can reveal actual agent behavior and support process optimization.

# 7. Conclusion

This paper presents a model-driven approach to Agentic Workflows in which BPMN Process Models are transformed into executable LangGraph/Pregel Workflows. Conventional activities, gateway conditions, process variables and human interactions remain part of the explicit process structure, while Agentic Tasks introduce LLM-based reasoning where deterministic execution is insufficient.

A central contribution is the use of Agentic Participants as reusable capability boundaries. Participants can be exposed as callable tools and may encapsulate direct agent behavior or an executable BPMN Process. This allows agent capabilities and process orchestration to be composed without requiring the calling agent to know their internal implementation.

Process Models and ontologies provide a complementary source of enterprise-specific context, helping agents interpret activities, business objects and organizational relationships. The same modeling environment can therefore connect the execution of work with knowledge about how an organization operates.

The underlying principle is that BPMN provides explicit process control, LangGraph/Pregel provides the execution runtime, and agents provide flexible reasoning within that structure. Rather than replacing established process models with autonomous agents, our approach makes them a foundation for orchestrating, composing and governing agentic capabilities.

Long-term goals are not to create gents that replace Process Models, but to create agents that:

**Understand, Execute and Optimize the Quality and Performance of Organizational Workflows**

# References

[1] Ait, A., Cánovas Izquierdo, J.L. and Cabot, J. (2025) ‘Towards Modeling Human-Agentic Collaborative Workflows: A BPMN Extension’, *arXiv preprint arXiv:2412.05958* [cs.SE]. Available at: https://arxiv.org/abs/2412.05958 (Accessed: 2 November 2026).

[2] Object Management Group (2014) *Business Process Model and Notation (BPMN), Version 2.0.2*. Needham: Object Management Group.

[3] LangChain (2026) *LangGraph Documentation: Graph API and Workflows and Agents*. Available at: https://docs.langchain.com/oss/python/langgraph/workflows-agents (Accessed: 2 November 2026).

[4] Calvanese, D. et al. (2026) ‘Agentic Business Process Management: A Research Manifesto’, *arXiv preprint arXiv:2603.18916* [cs.AI]. Available at: https://arxiv.org/abs/2603.18916 (Accessed: 2 November 2026).

[5] De Giacomo, G., Kampik, T., Kirchdorfer, L., Montali, M. and Weinhuber, C. (2026) ‘Formal Foundations of Agentic Business Process Management’, *arXiv preprint arXiv:2604.17347* [cs.AI]. Available at: https://arxiv.org/abs/2604.17347 (Accessed: 2 November 2026).

[6] Model Context Protocol (2025) *Specification: Server Overview*. Available at: https://modelcontextprotocol.io/specification/2025-06-18/server.

[7] Model Context Protocol (2025) *Specification: Server Tools*. Available at: https://modelcontextprotocol.io/specification/2025-11-25/server/tools.

[8] Alenezi, M. (2026) ‘Rethinking Software Engineering for Agentic AI Systems’, *arXiv preprint arXiv:2604.10599* [cs.SE]. Available at: https://arxiv.org/abs/2604.10599 (Accessed: 2 November 2026).

[9] TechByRahmat (2026) ‘LLM → RAG → AI Agents → Agentic AI: The Evolution Every Developer Should Understand’, *Medium*, 12 January. Available at: https://medium.com/skillstuff/llm-rag-ai-agents-agentic-ai-the-evolution-every-developer-should-understand-e6b2258d754c (Accessed: 2 November 2026).

[10] Level Up Coding (2026) ‘Context Engineering is Here... Because Prompts Alone Don't Scale’, *Level Up Coding*, 10 February. Available at: https://levelup.gitconnected.com/context-engineering-is-here-because-prompts-alone-dont-scale-ded6ea294b96 (Accessed: 2 November 2026).

[11] Santhanam, A. (2026) *Enterprise AI Infrastructure and Prompt Boundaries*. Industry Insights White Paper. F5.

[12] SemTalk, (2003) The business process modelling tool provided by Semtation GmbH https://www.semtalk.com